\documentclass{article}

\usepackage{microtype}
\usepackage{graphicx}
\usepackage{xspace}
\usepackage{subfigure}
\usepackage{booktabs} % for professional tables
\usepackage{tabularx}
\usepackage{algorithm2e}
\usepackage{xcolor}
\usepackage{multirow}
\usepackage{adjustbox}
\usepackage{fancyvrb}
\usepackage{tcolorbox}
\usepackage{}
\definecolor{googleblue}{HTML}{4285F4}
\definecolor{googlered}{HTML}{DB4437}
\definecolor{googlepurple}{HTML}{A142F4} % New purple color
\definecolor{googlegreen}{HTML}{0F9D58}

\usepackage{hyperref}

\usepackage[accepted]{icml2025}

\usepackage{amsmath}
\usepackage{amssymb}
\usepackage{mathtools}
\usepackage{amsthm}
\usepackage[percent]{overpic}

\usepackage[capitalize,noabbrev]{cleveref}

\theoremstyle{plain}
\newtheorem{theorem}{Theorem}[section]

\theoremstyle{definition}
\newtheorem{definition}[theorem]{Definition}

\theoremstyle{remark}
\newtheorem{remark}[theorem]{Remark}

\usepackage{etoolbox}
\BeforeBeginEnvironment{algorithm}{%
  \setlength{\intextsep}{8pt plus 2pt minus 2pt}%
  \setlength{\textfloatsep}{20pt plus 2pt minus 2pt}%
}

\newcounter{todocounter}
\usepackage{xcolor}
\definecolor{patblue}{HTML}{1F77B4}   % The exact hex code from your Python script
\definecolor{pgraphorange}{HTML}{FF7F0E}

\usepackage[textsize=tiny]{todonotes}

\usepackage{paralist}

\newtcolorbox{mybox}[1]{
  enhanced,
  breakable,
  colback=blue!5!white,
  colframe=blue!75!black,
  fonttitle=\mdseries,
  title=#1,
  boxrule=0.5pt,
  arc=3pt
}

\newcommand{\method}{\textsc{PaT}\xspace}

\icmltitlerunning{Sparse Personalized Text Generation with Multi-Trajectory Reasoning}

\begin{document}

\twocolumn[
\icmltitle{Sparse Personalized Text Generation with Multi-Trajectory Reasoning}

\icmlsetsymbol{equal}{*}

\begin{icmlauthorlist}
\icmlauthor{Bo Ni}{yyy}
\icmlauthor{Haowei Fu}{yyy}
\icmlauthor{Qinwen Ge}{yyy}
\icmlauthor{Franck Dernoncourt}{comp}
\icmlauthor{Samyadeep Basu}{comp}
\icmlauthor{Nedim Lipka}{comp}
\icmlauthor{Seunghyun Yoon}{comp}
\icmlauthor{Yu Wang}{sch}
\icmlauthor{Nesreen K. Ahmed}{comp}
\icmlauthor{Subhojyoti Mukherjee}{comp}
\icmlauthor{Puneet Mathur}{comp}
\icmlauthor{Ryan A. Rossi}{comp}
\icmlauthor{Tyler Derr}{yyy}

\end{icmlauthorlist}

\icmlaffiliation{yyy}{Department of Computer Science, Vanderbilt University, Nashville, Tennessee}
\icmlaffiliation{comp}{Adobe Research, San Jose, California}
\icmlaffiliation{sch}{University of Orgeon, Eugene, Oregon}

\icmlcorrespondingauthor{Tyler Derr}{tyler.derr@vanderbilt.edu}

\icmlkeywords{Personalization, RL}

\vskip 0.3in
]

\printAffiliationsAndNotice{\icmlEqualContribution} % 

\begin{abstract}
As Large Language Models (LLMs) advance, personalization has become a key mechanism for tailoring outputs to individual user needs. However, most existing methods rely heavily on dense interaction histories, making them ineffective in cold-start scenarios where such data is sparse or unavailable. While external signals (e.g., content of similar users) can offer a potential remedy, leveraging them effectively remains challenging: raw context is often noisy, and existing methods struggle to reason over heterogeneous data sources. To address these issues, we introduce \method (\textbf{P}ersonalization with \textbf{A}ligned \textbf{T}rajectories), a reasoning framework for cold-start LLM personalization. \method first retrieves information along two complementary trajectories: writing-style cues from stylistically similar users and topic-specific context from preference-aligned users. It then employs a reinforcement learning-based, iterative dual-reasoning mechanism that enables the LLM to jointly refine and integrate these signals. Experimental results across real-world personalization benchmarks show that \method consistently improves generation quality and alignment under sparse-data conditions, establishing a strong solution to the cold-start personalization problem. 
\end{abstract}

\section{Introduction}
\label{sec:introduction}
Large Language Model (LLM) personalization has received substantial attention due to its transformative potential in applications such as recommendation systems and conversational agents~\cite{zhang2025personalizationlargelanguagemodels, li2024personalizedlanguagemodelingpersonalized, au2025personalizedgraphbasedretrievallarge, kumar2024longlampbenchmarkpersonalizedlongform, salemi2024lamplargelanguagemodels}. By effectively leveraging personal context, personalized LLMs can generate content tailored to individual users, enhancing the user experience and fostering deeper engagement.

\begin{table}[t]
\renewcommand{\arraystretch}{0.5}
\setlength\tabcolsep{3.5pt}
\centering
\small
\scriptsize
\caption{Overall results.}
\label{tab:overall-results}
\vspace{2mm}
\begin{tabular}{cr cccc}
\toprule
% \textbf{Task} 
& \textbf{Metric} & \textbf{\method} (Ours) & \textbf{GraSPeR} & \textbf{PGraph} & \textbf{LaMP} \\
\midrule
\multirow{4}{*}{ \rotatebox{0}{\textcolor{googlegreen}{\textbf{\sc \bfseries Long Text Gen.}}} }
& R-1 $\uparrow$ & \textbf{0.233} & 0.195 & 0.213 & 0.182 \\[0.6ex]
& R-L $\uparrow$ & \textbf{0.162} & 0.147 & 0.146 & 0.124 \\[0.6ex]
& MET $\uparrow$ & \textbf{0.187} & 0.148 & 0.171 & 0.149 \\[0.6ex]
& LLM $\uparrow$ & \textbf{3.288} & 2.868 & 3.185 & 3.070 \\
\midrule
\multirow{4}{*}{\rotatebox{0}{\textcolor{googleblue}{\textbf{\sc \bfseries Short Text Gen.}}}}
& R-1 $\uparrow$ & \textbf{0.157} & 0.148 & 0.126 & 0.112 \\[0.6ex]
& R-L $\uparrow$ & \textbf{0.153} & 0.145 & 0.121 & 0.106 \\[0.6ex]
& MET $\uparrow$ & \textbf{0.128} & 0.124 & 0.124 & 0.117 \\[0.6ex]
& LLM $\uparrow$ & \textbf{3.541} & 3.240 & 3.396 & 3.225 \\
\bottomrule
\end{tabular}
\vskip -2ex
\end{table}

Existing strategies for LLM personalization heavily rely on the assumption of rich personal histories. For example, LaMP~\cite{salemi2024lamplargelanguagemodels} employs a Retrieval-Augmented Generation (RAG) framework to supply historical context to the prompt, while P-RLHF~\cite{li2024personalizedlanguagemodelingpersonalized} learns user embeddings derived from extensive past interactions. However, assuming the availability of rich histories is often not realistic in real-world settings, where thousands of new users emerge daily~\cite{zhang2025coldstartrecommendationeralarge}. Thus, a crucial challenge lies in addressing the cold-start problem—effectively personalizing for users with sparse or limited historical data.

To mitigate this data scarcity, recent work has incorporated auxiliary information to augment the limited user context. For example, PGraphRAG~\cite{au2025personalizedgraphbasedretrievallarge} leverages a user-topic graph to retrieve neighbor writings related to the generation target. Results demonstrate that incorporating this relevant context from peers can aid users with sparse histories. However, a key challenge remains: the retrieved context is often noisy and heterogeneous, encompassing signals ranging from the writing styles of similar users to distinct opinions on target topics. This complexity underscores the need for robust reasoning and a more sophisticated strategy to effectively integrate these heterogeneous signals into the generation process~\cite{salemi2025reasoningenhancedselftraininglongformpersonalized}.

To bridge this gap, we propose \method, a retrieval and reasoning framework for sparse LLM personalization. Instead of treating the retrieved content as a single, monolithic context, \method decomposes the personalization task into complementary trajectories: writing-style context from stylistically similar and topic-specific knowledge context from preference-aligned users. By employing a reinforcement-learning-based, iterative dual-reasoning mechanism, our approach enables the LLM to jointly refine and integrate these heterogeneous signals, filtering out noise while preserving critical personal markers. This allows the model to reason across multiple trajectories, making the final generation better aligned with user preferences, even under sparse conditions when the user’s own history is limited.

As shown in Table~\ref{tab:overall-results} and Figure~\ref{fig:teaser}, experiments on three real-world datasets demonstrate that \method outperforms state-of-the-art personalization baselines, gaining significant improvement on users with sparse history. In summary, our contributions are: % as follows:

\begin{figure}[t]
    \centering
    \vspace{1ex}
    \begin{overpic}[width=0.9\linewidth,
        trim=0 0 0 0, clip]{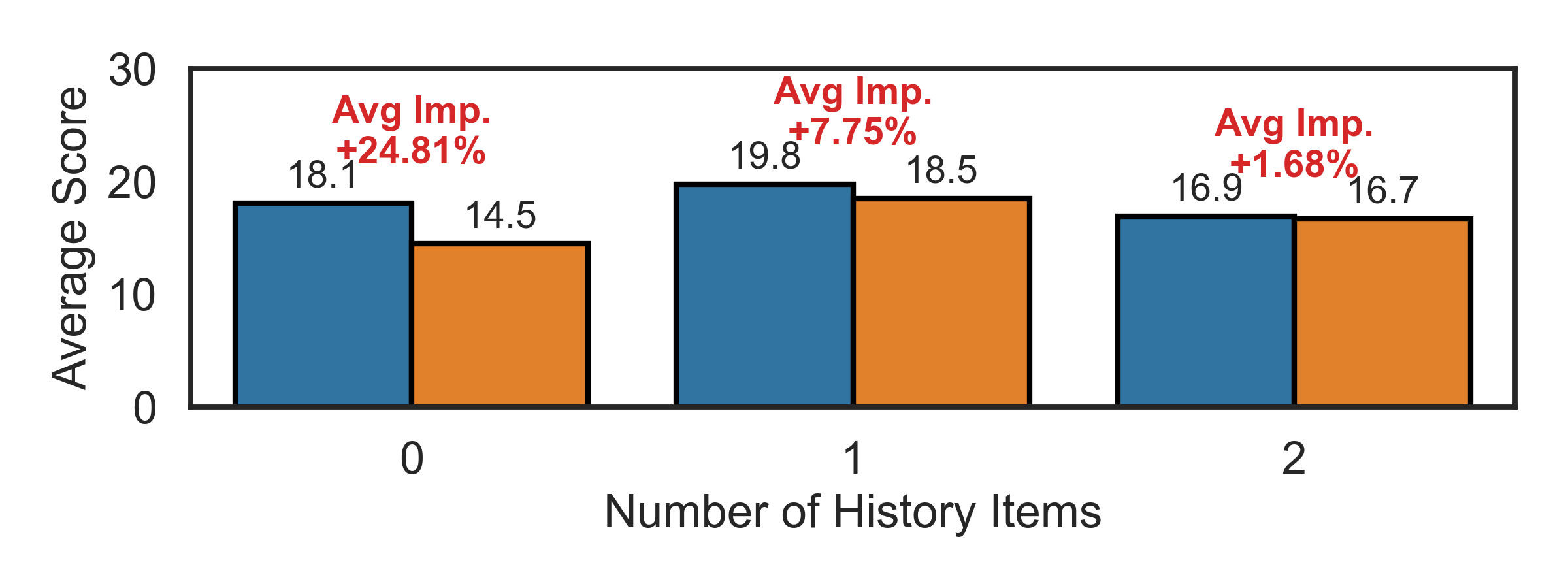}
        \put(31,35){\footnotesize
            \textcolor{patblue}{\rule[0.25ex]{0.6em}{0.6em}} \textbf{\small \method {\scriptsize ~(Ours)}}
            \hspace{0.5em}
            \textcolor{pgraphorange}{\rule[0.25ex]{0.6em}{0.6em}} \textbf{\small PGraph}
        }
    \end{overpic}
    \vskip -3ex
    \caption{Results comparing our approach across varying degrees of sparsity (amount of user history used).
    }
    \label{fig:teaser}
    \vskip -2ex
\end{figure}
\vspace{-2ex}
\begin{itemize} 
\item \textbf{Framework:} We introduce \method, a novel framework for cold-start LLM personalization that replaces simple context augmentation with multi-trajectory retrieval and reasoning. 
\vspace{-1ex}
\item \textbf{Context Augmentation:}  We propose a structured context augmentation strategy that decomposes retrieved heterogeneous information into complementary personalization signals.
\vspace{-1ex}
\item \textbf{Dual-Reasoning Mechanism:} We propose an iterative dual-reasoning paradigm that enables LLMs to synthesize heterogeneous style and preference signals, jointly optimizing the trajectories to improve personalization. 
\vspace{-1ex}
\item \textbf{Empirical Validation:} We conduct extensive experiments on three benchmark datasets, demonstrating that our approach improves generation quality and alignment in sparse-data scenarios. \end{itemize}

\section{Problem Definition}
\label{sec:problem}
In this section, we formally define the task and establish the notation used throughout this work. Let $\mathcal{U}$ be the set of users. The full notation is provided in Appendix~\ref{sec:notation}. Each user $u \in \mathcal{U}$ is associated with a personal history $\mathcal{H}_u = \{(x_{u,1}, y_{u,1}), \dots, (x_{u,n}, y_{u,n})\}$, where $x_{u,i}$ represents an input prompt and $y_{u,i}$ denotes the corresponding ground-truth text produced by the user. Additionally, let $\mathcal{A}_u$ represent any \textit{auxiliary information} for user $u$ (e.g., similar user profiles) that can supplement the user's context.

\begin{definition}[Personalized Text Generation]
Given a target user $u$, their historical context $\mathcal{H}_u$, auxiliary information $\mathcal{A}_u$, and a new target prompt $x_{target}$, the objective is to learn a parameterized function $f_\theta$ that generates a personalized output sequence $\hat{y}$:
\begin{equation}
    \hat{y} = f_\theta(x_{target}, \mathcal{H}_u, \mathcal{A}_u)
\end{equation}
\end{definition}

\begin{remark}[Cold-Start Personalization]
In this work, we specifically address the \textit{cold-start} scenario, where the user history $\mathcal{H}_u$ is sparse or limited in size (i.e., $|\mathcal{H}_u|$ is small). In such cases, $\mathcal{H}_u$ lacks sufficient signals for the model $f_\theta$ to effectively capture the user's specific writing style or preferences, often resulting in generic or misaligned generations. Consequently, the effective construction and utilization of auxiliary information $\mathcal{A}_u$ becomes crucial for mitigating this problem of data sparsity.
\end{remark}

\vspace{-1.25ex}
\section{Approach}
\label{sec:approach}
\begin{figure}
    \centering
    \includegraphics[width=\linewidth]{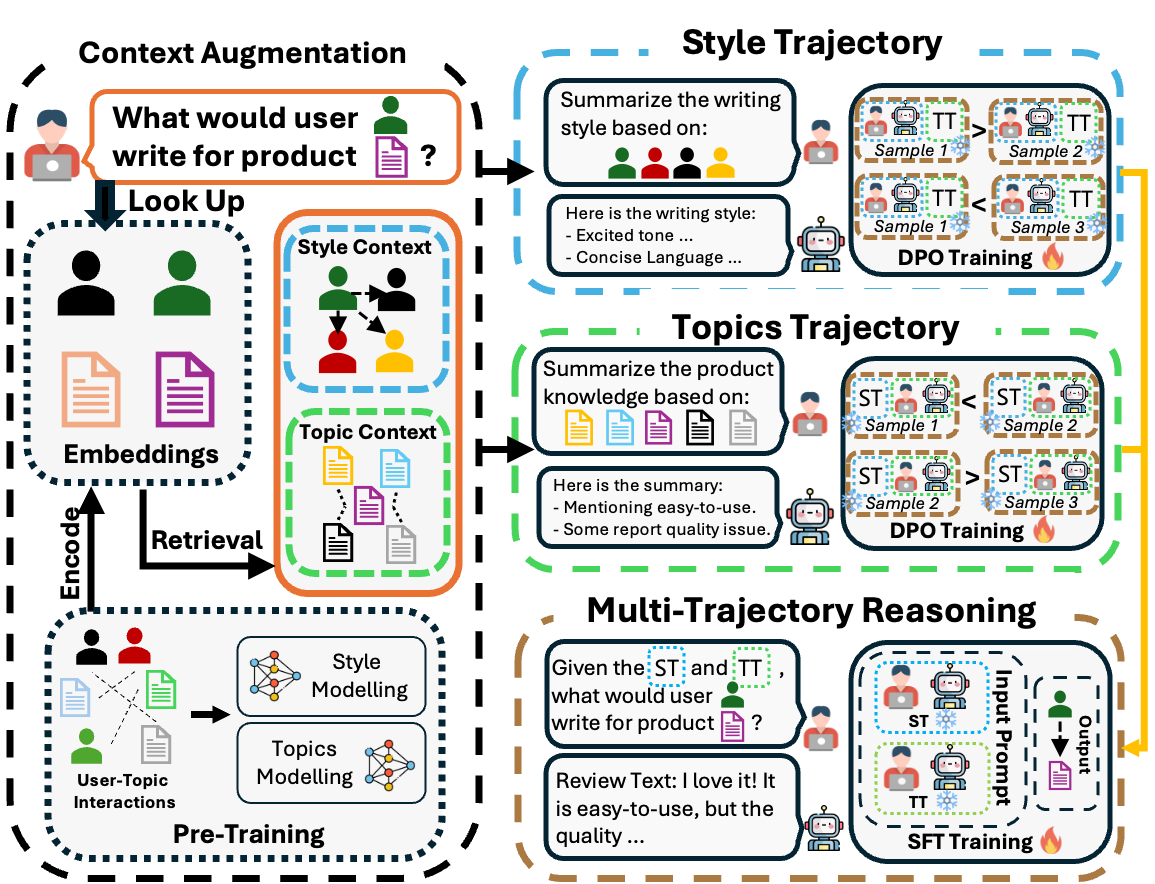}
    \vskip -2ex
    \caption{Overview of the proposed \method framework. The yellow arrow represents one training iteration.}
    \label{fig:overview}
    \vskip -2.5ex
\end{figure}

\vspace{-0.75ex}
In this section, we introduce \method, our proposed framework for cold-start personalization. As established in Section~\ref{sec:problem}, when the local history $\mathcal{H}_u$ is sparse, the \textit{construction} and \textit{utilization} of auxiliary information $\mathcal{A}_u$ become the key to generation quality. Shown in Figure~\ref{fig:overview}, on a high level, \method first identifies and augments the auxiliary information $\mathcal{A}_u$ with stylistically similar and preference-aligned neighbors. To effectively leverage $\mathcal{A}_u$ for personalization, \method then employs a multi-trajectory reasoning mechanism that enables the LLM to iteratively refine and integrate these heterogeneous signals into a personalized output. In the rest of this section, we will introduce them in more detail.

\vspace{-1ex}
\subsection{Context Augmentation}
\vspace{-0.75ex}
In cold-start scenarios where users have limited history, the sparse history $\mathcal{H}_u$ of user $u$ prevents the model from constructing effective user representations. To bridge this gap, leveraging auxiliary information from a neighborhood of related users has been widely adopted in recommendation systems~\cite{LIKA20142065} and recently discovered to improve personalization~\cite{au2025personalizedgraphbasedretrievallarge}.

To create the auxiliary information, we %first 
construct a user-topic bipartite graph $\mathcal{G} = (\mathcal{V}, \mathcal{E})$, %.
where $\mathcal{V} = \mathcal{U} \cup \mathcal{T}$ consists of users %nodes 
$u \in \mathcal{U}$ and entities %nodes 
$t \in \mathcal{T}$. For every historical entry $(x_{u,n}, y_{u,n}) \in \mathcal{H}_u$, we associate the text $y_{u,n}$ with a specific topic $t$. An edge $e_{u,t} \in \mathcal{E}$ exists if user $u$ has written about topic $t$. For example, if $y$ is user %written 
product reviews, then $t$ is the associated product; if $y$ is user %written 
emails, then $t$ is the email topic. % related to the email. 
Subsequently, $\mathcal{A}_u$ is defined by: 

\vspace{-3ex}
\begin{equation}
    \mathcal{A}_u = (\mathcal{C}^{style}_{u}, \mathcal{C}^{topic}_{u}) 
\end{equation}

\vspace{-2ex}
\noindent where $\mathcal{C}^{style}_{u}$ is the style context and $\mathcal{C}^{topic}_{u}$ is the topic knowledge context. We will discuss their construction separately in the rest of this subsection.

\subsubsection{Stylistic Context Extraction} The style context $\mathcal{C}^{style}_u$ aims to capture the linguistic patterns and tone of a given user. In cold-start scenarios, where $|\mathcal{H}_u|$ is small, the model cannot construct a reliable representation from the user's own history alone. To overcome this, we leverage graph learning to propagate stylistic cues across the user-topic bipartite graph $\mathcal{G}$. 

Let $\phi(\cdot)$ be a semantic encoder that maps text 
to a continuous embedding space. Since our goal is to propagate \emph{stylistic} information rather than topical semantics, we adopt a content-independent style encoder~\cite{wegmann2022authorjusttopiccontentindependent}. 

For each user node $u \in \mathcal{U}$, we construct the initial embedding by aggregating all prior texts written by $u$ in their historical profile. Concretely, this is obtained by mean-pooling the style embeddings of all associated texts:
\vspace{-0.75ex}
\begin{equation}
\mathbf{h}_u^{(0)} = \frac{1}{|\mathcal{H}_u|} \sum_{(x_{u,n}, y_{u,n}) \in \mathcal{H}_u} \phi(y_{u,n}).
\end{equation}

\vspace{-2.25ex}
Similarly, for each topic node $t \in \mathcal{T}$, we initialize its embedding by aggregating all texts associated with topic $t$ across different users. Let $\mathcal{R}_t$ denote the set of all texts written for topic $t$. The topic-level representation is obtained by mean-pooling the corresponding style embeddings:
\vspace{-0.75ex}
\begin{equation}
\mathbf{h}_t^{(0)} = \frac{1}{|\mathcal{R}_t|} \sum_{y \in \mathcal{R}_t} \phi(y).
\end{equation}

\vspace{-2.25ex}
To capture higher-order structural dependencies between users and entities, we apply a GraphSAGE encoder~\cite{hamilton2018inductiverepresentationlearninglarge} over the %bipartite 
graph $\mathcal{G}$ and obtain the final user embedding $\mathbf{h}_u = \mathbf{h}_u^{(L)}$ a after $L$ propagation layers.

We then retrieve the stylistically similar users for %user 
$u$ by computing cosine similarity 
of $\mathbf{h}_u$ with all other users:
\vspace{-1ex}
\begin{equation}
\mathcal{N}^{style}_u
= \operatorname*{arg\,topk_1}_{v \in \mathcal{U} \setminus \{u\}}
\frac{\mathbf{h}_u \cdot \mathbf{h}_v}{\|\mathbf{h}_u\| \, \|\mathbf{h}_v\|},
\end{equation}
where $\operatorname*{arg\,topk_1}$ returns the set of $k_1$ (hyperparameter) users with the highest cosine similarity.

The stylistic context is finally constructed by collecting historical texts from these neighbors:
\begin{equation}
\mathcal{C}^{style}_u = \{ y_{v,j} \mid v \in \mathcal{N}^{style}_u \},
\end{equation}
which provides the model with style-consistent exemplars independent of the target topic.

\subsubsection{Topic Knowledge Context Extraction}
While stylistic context captures \emph{how} a user tends to write, effective personalization also requires grounding the generation in \emph{what} should be written—namely, topic-specific knowledge and sentiment. The topic knowledge context $\mathcal{C}^{topic}_u$ is therefore designed to retrieve writings that are semantically aligned with the target topic.

Given a target topic $t_{\text{target}}$, we first identify all writings associated with the specific topic, forming an initial candidate set $\mathcal{Y}_\text{cand}(t_\text{target})$. However, exact topic matches can be sparse in real-world applications. To improve robustness, we introduce a semantic backoff mechanism at the topic level. Each topic is represented by a pooled embedding of its associated texts, and cosine similarity is used to retrieve a small set of semantically similar topics when insufficient exact matches are available.

To ensure the retrieved knowledge aligns with the target user's perspective, we rank the candidate writings based on the similarity between the target user $u$ and the author $v$ of each candidate text. Specifically, we prioritize authors whose overall preferences and behaviors are most aligned with the target user in the GraphSAGE embedding space:
\vspace{-1.5ex}
\begin{equation}
\mathcal{N}^{topic}_u = \operatorname*{arg\,topk_2}_{v \in \mathcal{U}_{\text{cand}}} \frac{\mathbf{h}_u \cdot \mathbf{h}_v}{\|\mathbf{h}_u\| \, \|\mathbf{h}_v\|},
\end{equation}

\vspace{-2.25ex}
where $\mathcal{U}_{\text{cand}}$ represents the set of users who have authored writings in the candidate topic set $\mathcal{Y}_\text{cand}$. $k_2$ is a hyperparameter denoting the maximum number of knowledge contexts retained. Formally, the topic knowledge context is defined as the set of writings regarding the target topic authored by these prioritized users:

\vspace{-2.75ex}
\begin{equation}
\mathcal{C}^{topic}_u = \{ y_{v,t_{\text{target}}} \mid v \in \mathcal{N}^{topic}_u \}.
\end{equation}

\subsection{Multi-Trajectory Reasoning}
\label{sec:multi-trajectory}
While the context augmentation stage constructs $\mathcal{A}_u = (\mathcal{C}^{style}_u, \mathcal{C}^{topic}_u)$, directly concatenating these raw neighbor texts into a single prompt is often noisy. For example, stylistic neighbors may contain irrelevant cues, and topic knowledge context may exhibit misaligned preferences. To robustly integrate these heterogeneous signals, \method employs a multi-trajectory reasoning mechanism that decomposes personalization into two complementary reasoning trajectories—(i) a \emph{style trajectory} that extracts user-level writing patterns from $\mathcal{C}^{style}_u$, and (ii) an \emph{topic knowledge trajectory} that extracts topic-specific knowledge and sentiment from $\mathcal{C}^{topic}_u$. The two trajectories are then fused to guide a downstream generation model.

\paragraph{Trajectory Reasoning Agents.}
We instantiate two reasoning agents parameterized by LLMs: a style reasoning agent $\pi_{\theta_s}$ and an topic knowledge reasoning agent $\pi_{\theta_t}$.
Given the retrieved neighbor texts, the style agent takes the context $\mathcal{C}_u^{style}$ together with the user's review history $\mathcal{H}_u$, and produces a style summary as the reasoning trajectory:
\begin{equation}
\label{eq:style}
s_u = \pi_{\theta_s}(\mathcal{P}_{style}(\mathcal{C}^{style}_u, \mathcal{H}_u))
\end{equation}
where $s_u$ denotes the summarized stylistic attributes (e.g., tone, phrasing, typical structure), and $\mathcal{P}_{style}$ denotes the prompt construction function that formats the style context and user history into the input sequence for the LLM. 

Similarly, the topic agent takes the topic knowledge context and produces a knowledge summary.
\begin{equation}
\label{eq:knowledge}
p_{u,t_{\text{target}}} = \pi_{\theta_t}(\mathcal{P}_{topic}(\mathcal{C}^{topic}_u))
\end{equation}
where $p_{u,t_{\text{target}}}$ captures topic context relevant to $t_{\text{target}}$ and $\mathcal{P}_{topic}$ denotes the prompt construction function that formulates the topic knowledge context into the LLM prompt.

\paragraph{Generation with Fused Trajectories.}
We then condition a generation model $\pi_{\theta_g}$ on both summarized trajectories (and optionally the original auxiliary texts) to produce the final output for the target topic:
\begin{equation}
\label{eq:gen}
\hat{y} = \pi_{\theta_g}(\mathcal{P}_{gen}(x_{target}, s_u, p_{u,t_\text{target}}, \mathcal{C}^{style}_u, \mathcal{C}^{topic}_u))
\end{equation}
where $\mathcal{P}_{gen}$ denotes the prompt construction function for final personalized text generation.

\begin{figure}[t]
    \centering
    \includegraphics[width=0.8\linewidth]{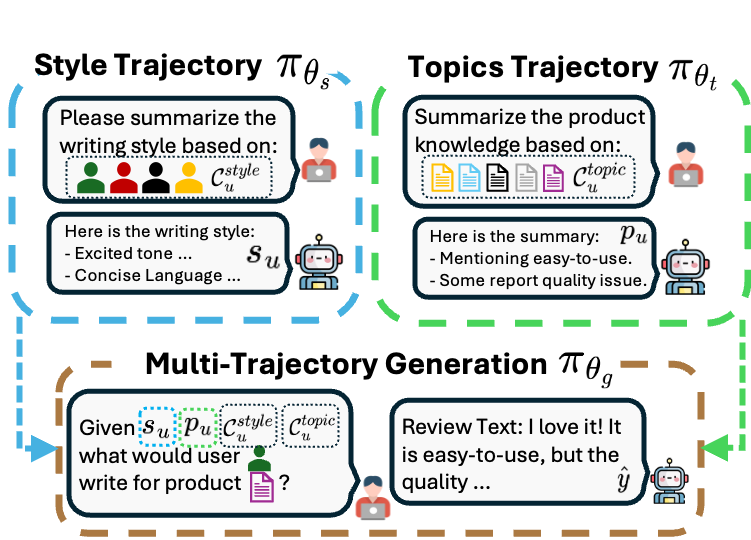}
    \vspace{-1ex}
    \caption{Computation graph of \method.}
    \label{fig:computation_graph}
    %\vspace{-1ex}
\end{figure}

\paragraph{Optimization via Differential Rewards.}
A key challenge is that the generation of the intermediate trajectory summaries cannot be directly supervised: there is no gold label for $s_u$ or $p_{u,t_{\text{target}}}$. To address this, we adopt an iterative self-improvement procedure based on differential reward signals~\cite{wang2021optimizingdatausagedifferentiable, li2025ragddroptimizingretrievalaugmentedgeneration}. At a high level, the trajectory reasoning agents are optimized by measuring how alternative trajectories affect the downstream generation quality, and we update the agent to prefer the better-performing trajectories. We visualize the method's computation graph in Figure~\ref{fig:computation_graph}. 

As shown in the figure, in one forward pass, $\pi_{\theta_g}$ takes the input from $\pi_{\theta_s}$ and $\pi_{\theta_t}$. We thus aim to optimize the trajectory agents $\pi_{\theta_s}$ and $\pi_{\theta_t}$ such that their intermediate outputs $s_u$ and $p_{u, t_\text{target}}$ induce higher-quality downstream generations $\hat{y}$. However, directly differentiating the generation loss with respect to the trajectory agents is intractable because of the discrete nature.

As a result, we leverage \emph{differential rewards} to assign credit to trajectory-level decisions.  Concretely, we sample multiple candidate summaries from each trajectory agent and roll them out through the generation model, obtaining a set of outputs whose relative task rewards reflect the quality of the underlying trajectories. Based on the relative rewards, we can use preference optimization~\cite{rafailov2024directpreferenceoptimizationlanguage} to update the underlying trajectory agent.   

Specifically, for the style reasoning agent, we sample $M_1$ candidate summaries:

\vspace{-2.25ex}
\begin{equation}
\big\{ s_u^{(m)} \big\}_{m=1}^{M_1} \sim \pi_{\theta_s}\!\left(\mathcal{P}_{\text{style}}(\mathcal{C}^{style}_u, \mathcal{H}_u)\right).
\end{equation}

\vspace{-1.5ex}
To evaluate the quality of each candidate summary, we fix the topic trajectory $p_u$ and roll out each $s_u^{(m)}$ through the downstream generation model, producing a generated output

\vspace{-3.5ex}
\begin{equation}
\label{eq:sample_style}
\hat{y}^{(m)} = \pi_{\theta_g}\!\left(
x_{\text{target}},\;
s_u^{(m)},\;
p_{u, t_\text{target}},\;
\mathcal{C}^{style}_u,\;
\mathcal{C}^{topic}_u
\right).
\end{equation}

\vspace{-2.25ex}
We then compute a task-specific reward by comparing the generated output $\hat{y}^{(m)}$ with the ground-truth target $y$:

\vspace{-2.25ex}
\begin{equation}
R^{(m)} = R\!\left(\hat{y}^{(m)}, y\right),
\end{equation}

\vspace{-2.25ex}
where $R(\cdot,\cdot)$ denotes the downstream evaluation function (e.g., ROUGE-based similarity for text generation).

Based on the relative rewards $\{ R^{(m)} \}$, we rank the sampled trajectory summaries and construct a preference dataset for the style reasoning agent. Concretely, for any pair of summaries $(s_u^{(i)}, s_u^{(j)})$ such that $R^{(i)} > R^{(j)}$, we treat $s_u^{(i)}$ as a preferred trajectory over $s_u^{(j)}$ and add the corresponding pairwise preference to the training set. These preference pairs are then used to update the style reasoning agent via Direct Preference Optimization (DPO).

Analogously, we optimize the topic reasoning agent by sampling $M_2$ candidate topic-aware summaries:
\begin{equation}
\{ p_{u, t_\text{target}}^{(m)} \}_{m=1}^{M_2} \sim \pi_{\theta_t}\!\left(\mathcal{P}_{topic}(\mathcal{C}^{topic}_u)\right).
\end{equation}
For each candidate $p_{u, t_\text{target}}^{(m)}$, we fix the style trajectory $s_u$ and roll out through the generation model, producing:
\begin{equation}
\label{eq:sample_knowledge}
\hat{y}^{(m)} = \pi_{\theta_g}\!\left(
x_{\text{target}},\;
s_u,\;
p_{u, t_\text{target}}^{(m)},\;
\mathcal{C}^{style}_u,\;
\mathcal{C}^{topic}_u
\right).
\end{equation}
We then compute the corresponding task reward $R^{(m)} = R(\hat{y}^{(m)}, y)$ by comparing the generated output with the ground-truth target. The knowledge trajectory reasoning agent is again optimized through DPO on the ranked trajectories based on the rewards.

\begin{algorithm}[t]
\label{code:pseudo}
\scriptsize 
\caption{\small{\method: Multi-Trajectory Self-Improvement with \\Differential Rewards for Personalized Text Generation}.}
\label{alg:method}
\DontPrintSemicolon
\KwIn{Training data $\mathcal{D}$; iterations $T$; samples $M_1,M_2$; neighbor sizes $k_1,k_2$}
\KwOut{Trajectory agents $\pi_{\theta_s}, \pi_{\theta_t}$ and generator $\pi_{\theta_g}$}

\BlankLine
\textbf{Precompute contexts.}
Build user--topic graph $\mathcal{G}$ and node embeddings; construct $\mathcal{C}^{style}_u$ (top-$k_1$ stylistic neighbors) and $\mathcal{C}^{topic}_u$ (top-$k_2$ topic-aligned neighbors)\;

\BlankLine
\For(){$t=1$ \KwTo $T$}{
% \For(\tcp*[f]{Iterative self-improvement}){$t=1$ \KwTo $T$}{
    \tcp{(1) Update style agent via differential rewards}
    Sample $\{s_u^{(m)}\}_{m=1}^{M_1}$ from $\pi_{\theta_s}(\mathcal{P}_{style}(\mathcal{C}^{style}_u,\mathcal{H}_u))$\;
    
    Roll out each $s_u^{(m)}$ through $\pi_{\theta_g}$ (fixing $p_u$) and compute rewards $R^{(m)}$\;
    
    Build pairwise preferences from reward ranking; update $\pi_{\theta_s}$ with DPO\;
    
    \vspace{0.25\baselineskip}
    \tcp{(2) Update topic agent via differential rewards}
    Sample $\{p_u^{(m)}\}_{m=1}^{M_2}$ from $\pi_{\theta_t}(\mathcal{P}_{topic}(\mathcal{C}^{topic}_u))$\;
    
    Roll out each $p_u^{(m)}$ through $\pi_{\theta_g}$ (fixing $s_u$) and compute rewards $R^{(m)}$\;
    
    Build pairwise preferences from reward ranking; update $\pi_{\theta_t}$ with DPO\;

    \vspace{0.25\baselineskip}
    \tcp{(3) Update generator on silver trajectories}
    Select silver summaries $(s_u^\star,p_u^\star)$ per instance; 
    
    SFT $\pi_{\theta_g}$ on $(x_{\text{target}}, s_u^\star, p_u^\star, y)$\;
}
\Return $\pi_{\theta_s}, \pi_{\theta_t}, \pi_{\theta_g}$\;
\end{algorithm}

Finally, to update the generation model $\pi_{\theta_g}$ to optimally utilize the refined trajectory summaries, we adopt a supervised fine-tuning (SFT) objective. Specifically, we construct training examples by conditioning the generator on the reasoning trajectories and the retrieved contexts, and train it to reproduce the ground-truth target output $y$. The generation model is optimized by minimizing the negative log-likelihood:

\vspace{-4.5ex}
\begin{equation}
\label{eq:sft}
\begin{aligned}
\min_{\theta_g} \;&
\mathbb{E}_{(x_{\text{target}}, s_u, p_{u, t_\text{target}}, y)}
\\
& \hspace{-2em}
\Big[
-\log \pi_{\theta_g}\!\Big(
y \mid
x_{\text{target}},\;
s_u,\;
p_{u, t_\text{target}},\;
\mathcal{C}^{style}_u,\;
\mathcal{C}^{topic}_u
\Big)
\Big].
\end{aligned}
\end{equation}

\vspace{-1.5ex}
This step is crucial in aligning the generator with the reasoning trajectories, further improving personalization.

\paragraph{Iterative Optimization.}
It is important to note that the trajectory reasoning agents and the generation model are inherently interdependent. Specifically, during trajectory sampling, the rollout in Eq.~\eqref{eq:sample_style} evaluates candidate style summaries by conditioning on the current generation model $\pi_{\theta_g}$ and a fixed topic summary produced by $\pi_{\theta_t}$, while Eq.~\eqref{eq:sample_knowledge} analogously evaluates topic-aware summaries in conjunction with $\pi_{\theta_g}$ and the style agent $\pi_{\theta_s}$. Conversely, the supervised fine-tuning objective in Eq.~\eqref{eq:sft} updates the generation model by conditioning its outputs on the trajectory summaries produced by $\pi_{\theta_s}$ and $\pi_{\theta_t}$. 
% As a result, improvements in any individual component directly influence the optimization landscape of the others.

To account for this interdependence, \method adopts an iterative optimization strategy that alternates between trajectory refinement and generator adaptation. At each iteration, we first optimize the style reasoning agent $\pi_{\theta_s}$ and the topic knowledge reasoning agent $\pi_{\theta_t}$ using differential reward signals derived from downstream generation quality, implemented via Direct Preference Optimization. Given the updated trajectory agents, we then update the generation model $\pi_{\theta_g}$ through supervised fine-tuning using the selected high-reward trajectory summaries.

This alternating optimization scheme progressively improves both the quality of intermediate reasoning trajectories and the generator’s ability to effectively exploit them. By propagating downstream task feedback to trajectory-level decisions through differential rewards, \method mitigates the absence of direct supervision for intermediate summaries and enables stable self-improvement, particularly in cold-start settings where user history is sparse and noisy. The pseudo-code of \method is provided in Algorithm~\ref{code:pseudo}.

\vspace{-1ex}
\section{Experiments}
\label{sec:experiments}
\vspace{-0.75ex}
In this section, we conduct extensive experiments to evaluate the performance of \method on the personalized text generation. 

\vspace{-0.75ex}
\subsection{Datasets and Metrics}
\vspace{-0.75ex}
Our experimental design follows established evaluation protocols in prior studies~\citep{au2025personalizedgraphbasedretrievallarge} and is conducted on three commonly used benchmarks: Amazon Reviews~\citep{ni-mcauley-2018-personalized}, Hotel Reviews~\citep{kanouchi-etal-2020-may}, and Stylized Feedback~\citep{alhafni-etal-2024-personalized}. All three datasets consist of large-scale human-authored texts and exhibit a pronounced cold-start setting, with over 95\% of users having fewer than two historical writings. The dataset statistics are presented in Appendix~\ref{app:dataset}. We evaluate performance on long-form text generation and short-form text generation. In the long-form generation task, the model is conditioned on a title prompt, while the short-form generation task involves summarizing a given paragraph.

For both generation tasks, we report standard lexical similarity metrics, including ROUGE-1, ROUGE-L, and METEOR, consistent with prior work~\citep{au2025personalizedgraphbasedretrievallarge, kumar2024longlampbenchmarkpersonalizedlongform, salemi2024lamplargelanguagemodels}. Since such metrics primarily capture surface-level overlap, we additionally employ an LLM-as-a-Judge framework~\citep{liu-etal-2023-g}, in which a high-capacity language model performs pairwise evaluations of personalization quality. Details of the LLM-as-a-Judge protocol are provided in Appendix~\ref{app:lj}. A comprehensive description of the evaluation metrics and experimental configuration is included in Appendix~\ref{app:exp-set-up}.

\begin{table}[t]
\renewcommand{\arraystretch}{0.4}
\setlength\tabcolsep{1.25pt}
\centering
\small
\caption{
Results on the Amazon Review Generation Benchmark.}
\label{tab:amazon_results}
\vspace{2mm}
\begin{tabular}{ccccccc}
\toprule
\textbf{Task} & \textbf{Metric} & \textbf{LLM} & 
\textbf{\method} & 
\textbf{GraSPeR} & 
\textbf{PGraph} & \textbf{LaMP} \\
\midrule
\multirow{9}{*}{ \rotatebox{90}{\textcolor{googlegreen}{\textbf{\sc \bfseries Long Text Gen.}}} }
& \multirow{2}{*}{R-1 $\uparrow$}  & \textit{Qwen3} & \textbf{0.210} & 0.168 & 0.183 & 0.168 \\[0.6ex]
&                           & \textit{LlaMA3}     & \textbf{0.197} & 0.167 & 0.152 & 0.161 \\
\cmidrule(l){2-7}
& \multirow{2}{*}{R-L $\uparrow$}  & \textit{Qwen3}             & \textbf{0.159} & 0.129 & 0.132 & 0.122 \\[0.6ex]
&              & \textit{LlaMA3}               & \textbf{0.151} & 0.141 & 0.133 & 0.114 \\
\cmidrule(l){2-7}
& \multirow{2}{*}{MET $\uparrow$}  & \textit{Qwen3}             & \textbf{0.176} & 0.150 & 0.177 & 0.158 \\[0.6ex]
&                           & \textit{LlaMA3}               & \textbf{0.158} & 0.124 & 0.109 & 0.149 \\
\cmidrule(l){2-7}
& \multirow{2}{*}{LLM $\uparrow$}  & \textit{Qwen3}  & 3.432 & 2.961  & \textbf{3.725} & 3.353 \\[0.6ex]
&              & 

\textit{LlaMA3} & 3.100 & 3.006 & 3.399 & \textbf{3.701} \\
% \textit{LlaMA3} & \textbf{0.092} & 0.018 & 0.015 & 0.011 \\
\midrule
\multirow{9}{*}{\rotatebox{90}{\textcolor{googleblue}{\textbf{\sc \bfseries Short Text Gen.}}}}
& \multirow{2}{*}{R-1 $\uparrow$}  & \textit{Qwen3}             & \textbf{0.177} & 0.158 & 0.124  & 0.107 \\[0.6ex]
&                           & \textit{LlaMA3}               & \textbf{0.150} & 0.132 & 0.120 & 0.115 \\
\cmidrule(l){2-7}
& \multirow{2}{*}{R-L $\uparrow$} & \textit{Qwen3}             & \textbf{0.171} & 0.155 & 0.120  & 0.102 \\[0.6ex]
&                           & \textit{LlaMA3}               & \textbf{0.147} & 0.129 & 0.115 & 0.109 \\
\cmidrule(l){2-7}
& \multirow{2}{*}{MET $\uparrow$} & \textit{Qwen3}             & \textbf{0.155} & 0.134 & 0.132  & 0.107 \\[0.6ex]
&                           & \textit{LlaMA3}               & 0.116 & \textbf{0.128} & 0.121 & 0.124 \\
\cmidrule(l){2-7}
& \multirow{2}{*}{LLM $\uparrow$}   & \textit{Qwen3}  & \textbf{3.744} & 3.388 & 3.722 & 3.460 \\[0.6ex]
&              
& \textit{LlaMA3}               & \textbf{3.261} & 3.212 & 3.141 & 3.081 \\
% & \textit{LlaMA3}               & 0.075 & 0.111 & 0.052 & 0.041 \\
\bottomrule
\end{tabular}
\vskip -2ex
\end{table}

\vspace{-0.75ex}
\subsection{Baselines}
\vspace{-0.75ex}
We benchmark against three state-of-the-art personalization baselines. \textbf{LaMP}~\citep{salemi2024lamplargelanguagemodels} conditions on a user's past writing via prompts but uses no graph learning or reasoning. \textbf{PGraphRAG}~\citep{au2025personalizedgraphbasedretrievallarge} (i.e., PGraph) employs graph-based retrieval augmented generation with BM25 for personalization but lacks reasoning or fine-tuning. \textbf{GraSPeR}~\citep{ni2026grasper} augments sparse users' context with graph learning, and leverages reasoning to further refine the outputs. Expanded baseline descriptions can be found in Appendix~\ref{app:exp-set-up}. We evaluate all baselines with Qwen3-4B-Instruct~\cite{bai2023qwentechnicalreport} and Llama3.2-3B-Instruct~\cite{grattafiori2024llama3herdmodels}.

\subsection{Main Results}

We present the main experimental results in Table~\ref{tab:amazon_results}, Table~\ref{tab:hotel_results}, and Table~\ref{tab:gap_results} for the Amazon Review, Hotel Experience, and Stylized Feedback benchmarks, respectively. Across all datasets and tasks, \method consistently achieves strong performance compared to existing personalization baselines.

On long-text generation tasks, \method consistently outperforms all baselines across lexical metrics (ROUGE-1, ROUGE-L, and METEOR), indicating more faithful personalized generation. This improvement can be attributed to the fact that long-text generation inherently requires multi-step reasoning and sustained contextual coherence. By explicitly constructing and refining trajectory-level representations, \method is able to aggregate and organize user- and topic-specific signals over multiple reasoning steps, leading to more informative and better-aligned generation outputs.

For short-text generation tasks, \method remains competitive and generally achieves the best or near-best performance. However, compared to the long-text generation results, the relative improvements are smaller. This suggests that short-text generation is less sensitive to complex reasoning trajectories, as the limited output space reduces the need for deep preference aggregation. In such settings, surface-level semantic matching or single-hop personalization signals are often sufficient, which narrows the performance gap between \method and existing baselines.

Beyond surface-level metrics, \method demonstrates clear advantages under LLM-as-a-Judge evaluations. Across all three datasets, \method achieves top (or near-top) LLM-as-a-Judge scores. %, with the greatest improvements appearing in several configurations. 
This suggests %that 
the improvements of %brought by 
\method extend beyond n-gram overlap and translate into higher-quality, more preference-aligned outputs as judged by strong language models. These improvements are particularly meaningful in short-text generation, where capturing nuanced user preferences is critical, and we observe notable gains.

\begin{table}[t]
\renewcommand{\arraystretch}{0.4}
\setlength\tabcolsep{1.25pt}
\centering
\small
\caption{Results on the Hotel Experience Benchmark.}
\label{tab:hotel_results}
\vspace{2mm}
\begin{tabular}{ccccccc}
\toprule
\textbf{Task} & \textbf{Metric} & \textbf{LLM} & 
\textbf{\method} & 
\textbf{GraSPeR} & 
\textbf{PGraph} & \textbf{LaMP} \\
\midrule
\multirow{8}{*}{ \rotatebox{90}{\textcolor{googlegreen}{\textbf{\sc \bfseries Long Text Gen.}}} }
& \multirow{2}{*}{R-1 $\uparrow$}  & \textit{Qwen3}  & 0.240 & 0.234 & \textbf{0.254} & 0.178 \\[0.6ex]
&                           & \textit{LlaMA3}      & \textbf{0.277} & 0.211 & 0.260 & 0.224 \\
\cmidrule(l){2-7}
& \multirow{2}{*}{R-L $\uparrow$}  & \textit{Qwen3}             & 0.142 & \textbf{0.156} & 0.148 & 0.117 \\[0.6ex]
&              & \textit{LlaMA3}               & \textbf{0.167} & 0.148 & 0.155 & 0.141 \\
\cmidrule(l){2-7}
& \multirow{2}{*}{MET $\uparrow$}  & \textit{Qwen3}             & \textbf{0.194} & 0.163 & 0.193 & 0.114 \\[0.6ex]
&                           & \textit{LlaMA3}               & \textbf{0.204} & 0.152 & 0.182 & 0.148 \\
\cmidrule(l){2-7}
& \multirow{2}{*}{LLM $\uparrow$}  & \textit{Qwen3}  & \textbf{3.438} & 2.75  & 3.174 & 2.733 \\[0.6ex]
&              & 

\textit{LlaMA3} & \textbf{3.231} & 2.520 & 1.838 & 2.651 \\
% \textit{LlaMA3} & \textbf{0.094} & 0.041 & 0.052 & 0.039 \\
\midrule
\multirow{8}{*}{\rotatebox{90}{\textcolor{googleblue}{\textbf{\sc \bfseries Short Text Gen.}}}}
& \multirow{2}{*}{R-1 $\uparrow$}  & \textit{Qwen3}             & 0.133 & \textbf{0.150} & 0.122  & 0.112 \\[0.6ex]
&                           & \textit{LlaMA3}               & \textbf{0.126} & 0.120 & 0.125 & 0.104 \\
\cmidrule(l){2-7}
& \multirow{2}{*}{R-L $\uparrow$} & \textit{Qwen3}             & 0.123 & \textbf{0.145} & 0.114  & 0.104 \\[0.6ex]
&                           & \textit{LlaMA3}               & \textbf{0.121} & 0.117 & 0.116 & 0.096 \\
\cmidrule(l){2-7}
& \multirow{2}{*}{MET $\uparrow$} & \textit{Qwen3}             & \textbf{0.118} & \textbf{0.118} & 0.106  & 0.096 \\[0.6ex]
&                           & \textit{LlaMA3}               & 0.085 & 0.108 & \textbf{0.111} & 0.100 \\
\cmidrule(l){2-7}
& \multirow{2}{*}{LLM $\uparrow$}   & \textit{Qwen3}  & \textbf{3.739} & 3.282 & 3.651 & 3.512 \\[0.6ex]
&              
& \textit{LlaMA3}               & \textbf{3.462} & 3.087 & 3.196 & 3.051 \\
% & \textit{LlaMA3}               & \textbf{0.071} & 0.058 & 0.064 & 0.051 \\
\bottomrule
\end{tabular}
%\vskip -0.5ex
\end{table}

\begin{figure}[t]
    \centering
    \includegraphics[width=0.75\linewidth]{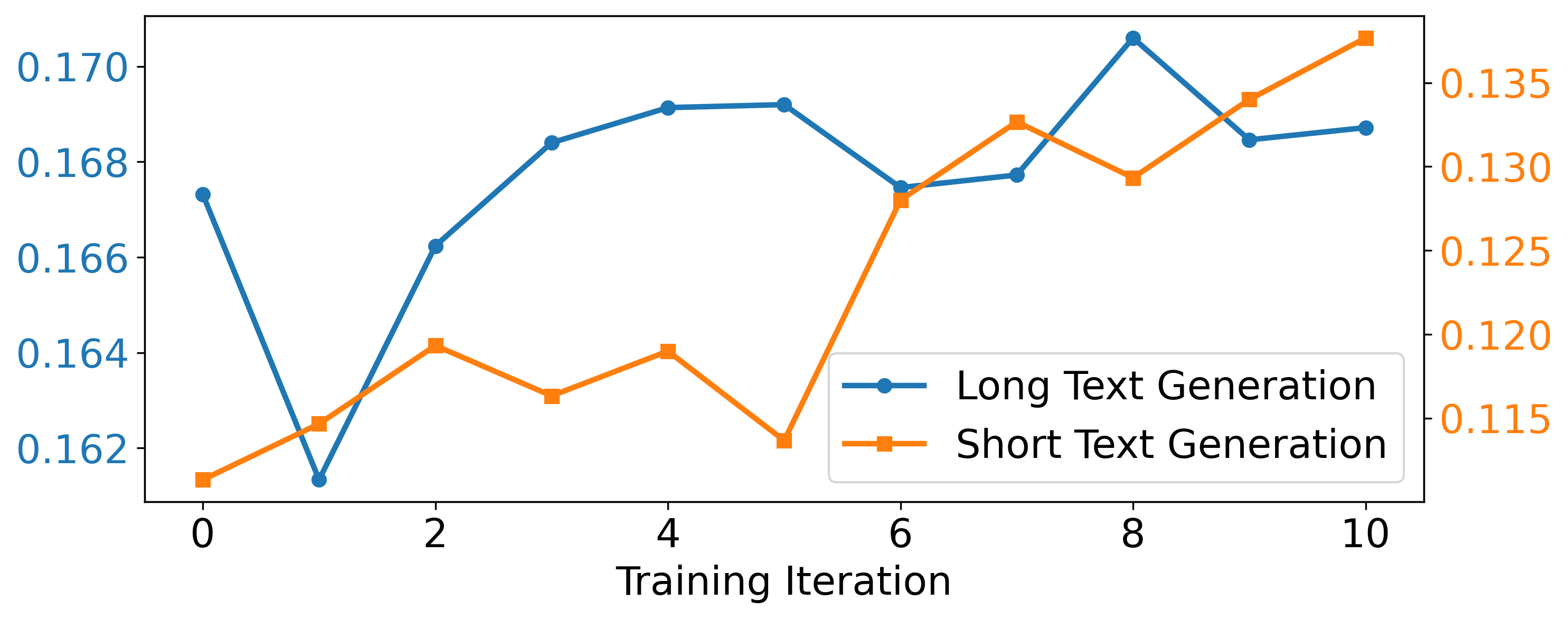}
    \vspace{-2ex}
    \caption{Text generation performance as iteration number increases on the Amazon Review dataset.}
    % \ryan{multiple runs can help smooth this out if time permits}}
    \label{fig:convergence}
    \vspace{-3ex}
\end{figure}

Moreover, the consistent gains across datasets highlight the robustness of \method under diverse personalization scenarios and content domains, suggesting that structured trajectory refinement is a crucial component for reliable personalized generation, particularly for tasks requiring deeper reasoning and longer outputs.

\begin{table}[t]
\renewcommand{\arraystretch}{0.4}
\setlength\tabcolsep{1.25pt}
\centering
\small
\caption{Results on the Stylized Feedback Generation Benchmark.}
\label{tab:gap_results}
\vspace{2mm}
\begin{tabular}{ccccccc}
\toprule
\textbf{Task} & \textbf{Metric} & \textbf{LLM} & 
\textbf{\method} & 
\textbf{GraSPeR} & 
\textbf{PGraph} & \textbf{LaMP} \\
\midrule
\multirow{8}{*}{ \rotatebox{90}{\textcolor{googlegreen}{\textbf{\sc \bfseries Long Text Gen.}}} }
& \multirow{2}{*}{R-1 $\uparrow$}  & \textit{Qwen3}  & \textbf{0.242} & 0.196 & 0.213 & 0.179 \\[0.6ex]
&                           & \textit{LlaMA3}      & \textbf{0.231} & 0.191 & 0.211 & 0.181 \\
\cmidrule(l){2-7}
& \multirow{2}{*}{R-L $\uparrow$}  & \textit{Qwen3}             & \textbf{0.179} & 0.149 & 0.153 & 0.124 \\[0.6ex]
&              & \textit{LlaMA3}               & \textbf{0.175} & 0.154 & 0.152 & 0.122 \\
\cmidrule(l){2-7}
& \multirow{2}{*}{MET $\uparrow$}  & \textit{Qwen3}             & \textbf{0.210} & 0.160 & 0.191 & 0.161 \\[0.6ex]
&                           & \textit{LlaMA3}               & \textbf{0.179} & 0.134 & 0.175 & 0.164 \\
\cmidrule(l){2-7}
& \multirow{2}{*}{LLM $\uparrow$}  & \textit{Qwen3}  & 3.357 & 3.040  & \textbf{3.685} & 3.107 \\[0.6ex]
&              & 
\textit{LlaMA3} & 3.171 & 2.930 & \textbf{3.290} & 2.873 \\
% \textit{LlaMA3} & \textbf{0.087} & 0.054 & 0.061 & 2.896 \\
\midrule
\multirow{8}{*}{\rotatebox{90}{\textcolor{googleblue}{\textbf{\sc \bfseries Short Text Gen.}}}}
& \multirow{2}{*}{R-1 $\uparrow$}  & \textit{Qwen3}             & \textbf{0.191} & 0.170 & 0.127  & 0.115 \\[0.6ex]
&                           & \textit{LlaMA3}               & \textbf{0.166} & 0.157 & 0.139 & 0.121 \\
\cmidrule(l){2-7}
& \multirow{2}{*}{R-L $\uparrow$} & \textit{Qwen3}             & \textbf{0.189} & 0.168 & 0.122  & 0.110 \\[0.6ex]
&                           & \textit{LlaMA3}               & \textbf{0.163} & 0.155 & 0.134 & 0.116 \\
\cmidrule(l){2-7}
& \multirow{2}{*}{MET $\uparrow$} & \textit{Qwen3}             & \textbf{0.174} & 0.127 & 0.134  & 0.135 \\[0.6ex]
&                           & \textit{LlaMA3}               & 0.116 & 0.127 & \textbf{0.139} & 0.136 \\
\cmidrule(l){2-7}
& \multirow{2}{*}{LLM $\uparrow$}   & \textit{Qwen3}  & \textbf{3.716} & 3.292 & 3.586 & 3.356 \\[0.6ex]
&              & 
\textit{LlaMA3}               & \textbf{3.325} & 3.175 & 3.076 & 2.888 \\
% \textit{LlaMA3}               & \textbf{0.072} & 0.061 & 0.066 & 3.423 \\
\bottomrule
\end{tabular}
\vspace{-0.5em}
\end{table}
\begin{table}[t]
\renewcommand{\arraystretch}{0.4}
\centering
\scriptsize
\caption{Comparison between PAT and PGraph with Average Improvement on the Short Text Generation for the Amazon dataset.
% \td{lets think about 3+}
% \bo{GraSPeR results}.
}
\label{tab:comparison_updated}
\vspace{2mm}
\begin{tabular}{llcccc}
\toprule
\# History & Method  & ROUGE-1 & ROUGE-L & METEOR &  $\Delta$\% \\
\midrule
\multirow{2}{*}{0}  & PAT     & 20.73 & 15.82 & 17.69 & \multirow{2}{*}{+24.81\%} \\[0.6ex]
                    & PGraph  & 16.23 & 12.53 & 14.69 & \\
\midrule
\multirow{2}{*}{1}  & PAT     & 23.10 & 17.63 & 18.58 & \multirow{2}{*}{+7.75\%} \\[0.6ex]
                    & PGraph  & 21.39 & 16.30 & 17.67 & \\
\midrule
\multirow{2}{*}{2}  & PAT     & 21.61 & 14.23 & 14.88 & \multirow{2}{*}{+1.68\%} \\[0.6ex]
                    & PGraph  & 20.46 & 12.98 & 16.56 & \\
% \midrule
% \multirow{2}{*}{3+} & PAT     & 24.59 & 16.34 & 16.98 & \multirow{2}{*}{-7.56\%} \\
%                     & PGraph  & 26.05 & 16.55 & 20.16 & \\
\bottomrule
\end{tabular}
\end{table}
\subsection{Model Convergence}

Beyond static performance, we investigate the learning dynamics and stability of our approach. To this end, we track the convergence behavior of \method by plotting the average performance metrics for both text and title generation on the Amazon Review dataset across training iterations, as illustrated in Figure~\ref{fig:convergence}.

We observe that as the iteration count increases, the model performance exhibits a general upward trend across all primary metrics. Specifically, the metrics show a steep initial improvement followed by a gradual stabilization, suggesting that \method efficiently explores the solution space before converging to a high-performing local optimum. This steady convergence behavior on both tasks demonstrates the robustness of the \method architecture and validates its effectiveness in iteratively refining generation quality.

\paragraph{Note on the training epochs.}
To ensure that these performance gains are a result of the iterative refinement process rather than simply an artifact of increased training duration (i.e., more total epochs), we implement a strict early-stopping criterion. Specifically, we set a limit of 50 steps for both the Supervised Fine-Tuning (SFT) and Direct Preference Optimization (DPO) phases within each iteration. This ensures that each individual iteration reaches a point of local convergence independently. Consequently, the observed improvements across successive iterations can be attributed to the quality of the generated trajectories and the iterative feedback loop, rather than cumulative training time.

\subsection{Performance Analysis by History Length}

We further provide a granular evaluation of \method across different user history lengths in Table~\ref{tab:comparison_updated}. The results highlight that \method is particularly effective at addressing the ``cold-start" problem (i.e., users with sparse interaction data). For users with zero history, \method outperforms the PGraph baseline by an average of $24.81\%$ across metrics.

As history length increases, the relative performance gain diminishes. This suggests that while \method excels at extracting signal from minimal data, existing methods may better capture long-term dependencies as the history becomes richer. In production, these findings indicate that a stratified deployment strategy where users are routed to different models based on their history density could potentially further optimize overall system performance.

\section{Ablation Study}
\begin{table}[t]
\centering
\footnotesize 
\caption{Ablation study of \method highlighting the contribution of style and product trajectories. We report ROUGE-1 (R-1), ROUGE-L (R-L), METEOR, and LLM-as-a-Judge (LLM) scores.}
\label{tab:ablation}
\vspace{2mm}
\begin{tabularx}{\columnwidth}{Xcccc}
\toprule
\textbf{Variant} & \textbf{R-1} & \textbf{R-L} & \textbf{MET} & \textbf{LLM} \\
\midrule
\method & \textbf{0.18} & \textbf{0.17} & \textbf{0.16} & \textbf{3.75} \\
w/o Style Trajectory & 0.16 & 0.16 & 0.15 & 3.54 \\
w/o Knowledge Trajectory & 0.17 & 0.16 & 0.16 & 3.58 \\
w/o Both Trajectories & 0.13 & 0.12 & 0.12 & 3.24 \\
w/o Training (Zero-shot) & 0.12 & 0.11 & 0.13 & 3.35 \\
\bottomrule
\end{tabularx}
\vspace{-0.5ex}
\end{table}

\label{sec:ablation}
In this section, we conduct an extensive ablation study to investigate the contribution of \method's core components to its overall performance. Our analysis is driven by two central research questions: (1) To what extent do the style and topic knowledge reasoning trajectories contribute to effective personalization? and (2) How does the iterative DPO and SFT training procedure improve upon a zero-shot retrieval-augmented baseline? Table~\ref{tab:ablation} presents the results measured by ROUGE-1 (R-1), ROUGE-L (R-L), METEOR (MET), and LLM-as-a-Judge (LLM) scores.

\begin{table*}[t]
    \centering
    \scriptsize
        \caption{Case Study. Orange text shows traces that are too general. Red text show traces that do not help during generation. Green text shows correct reasoning traces that directly contribute to the correct personalization.}
        \vspace{2ex}
    % REDUCED FROM 1.5 to 1.1
    \renewcommand{\arraystretch}{1.1} 
    \label{tab:case-study}
    
    % Added >{\raggedright\arraybackslash} to X columns for tighter text wrapping
    \begin{tabularx}{\linewidth}{ @{} l >{\raggedright\arraybackslash}X >{\raggedright\arraybackslash}X @{} }
        \toprule
        & \multicolumn{1}{c}{\textcolor{googlered}{\textsc{Iteration 0 (Baseline)}}} 
        & \multicolumn{1}{c}{\textcolor{googlegreen}{\textsc{Iteration 4 (Refined)}}} \\
        \midrule % Added midrule here to separate header
        
        \textsc{Ground Truth}
        & \multicolumn{2}{c}{ % Changed to 'c' to use natural width, or keep p{...} if preferred
            \textit{Target Review:} "Love this! Beautiful red. Stays on all day!"
          } \\ 
        \midrule

        \textsc{Style Trace} 
        & \textcolor{orange}{Informal tone}; \textcolor{orange}{Positive language}; Focus on product experiences and \textcolor{googlered}{personal preferences}; Emphasis on \textcolor{orange}{product quality}.
        & \textcolor{googlegreen}{Clumsy tone} and \textcolor{googlegreen}{short sentences}; Use of \textcolor{googlegreen}{emphatic language (e.g. Wasted my money!)}; \textcolor{googlegreen}{Use of colloquial expression}. \\
        % REMOVED \addlinespace
        
        \textsc{Topic Trace} 
        & \textcolor{orange}{Brand: Maybelline}, \textcolor{googlered}{Quality: Poor}, \textcolor{googlegreen}{Color: Red}, \textcolor{googlegreen}{Staying Power: Good}.
        & \textcolor{googlegreen}{The product is a lipstick with a vibrant red color that stays in place well.} \textcolor{orange}{Considered affordable.} \\
        % REMOVED \addlinespace
        
        \textsc{Final Text} 
        & \textcolor{googlegreen}{I love this lipstick}. It is a \textcolor{googlegreen}{great color} and \textcolor{orange}{it is so easy to apply}. It is long lasting and \textcolor{googlered}{it does not bleed. I will definitely be buying more.}
        & I \textcolor{googlegreen}{love the color} and the \textcolor{googlegreen}{long lasting finish!}
        \\
        
        \bottomrule
    \end{tabularx}
    %\vskip -1ex

\end{table*}
\subsection{Impact of Multi-Trajectory Reasoning}
To evaluate the model architecture of our framework, we compare the full \method against several variants that selectively remove the reasoning trajectories. As shown in Table~\ref{tab:ablation}, removing either the topic knowledge trajectory (\textit{w/o Knowledge Trajectory}) or the style trajectory (\textit{w/o Style Trajectory}) leads to a consistent decline in performance across all metrics. Furthermore, the most significant degradation occurs in the case when both trajectories are excluded, where the model is provided only with raw retrieved information $\mathcal{A}_u$. These results consolidate our hypothesis that both trajectories are essential; while removing one hampers performance, removing both renders the worst performance, confirming that intermediate reasoning is vital for synthesizing noisy and potentially misaligned signals from the neighborhood.

\subsection{Impact of Iterative Training}
To answer our second research question, we evaluate \method in a zero-shot configuration, where we utilize the proposed retrieval and reasoning architecture but replace the fine-tuned agents ($\pi_{\theta_s}, \pi_{\theta_t}, \pi_{\theta_g}$) with their base LLM counterparts.

As illustrated in Table~\ref{tab:ablation}, the ``w/o Training (Zero-shot)'' variant yields the lowest performance across the board. While the retrieval-augmented framework provides the model with the necessary contextual knowledge, the base LLM lacks the alignment required to prioritize the specific stylistic cues and domain knowledge that lead to high-quality personalization. This substantial gap highlights the effectiveness of our iterative optimization strategy: by leveraging differential rewards and DPO, \method learns to prefer reasoning trajectories that improve downstream generation quality.

\vspace{-0.5ex}
\section{Case Study}
To validate the efficacy of our multi-trajectory reasoning, we present a qualitative analysis of the intermediate reasoning traces and generated outputs in Table~\ref{tab:case-study}. In Iteration 0, the model fails to filter noise from the raw retrieved context; the Style Trace remains generic, while the Topic Trace hallucinates negative attributes. Consequently, the baseline generation resorts to a safe, generic review that lacks user-specific character. In contrast, after optimization with differential data rewards, the Topic Knowledge Trajectory successfully isolates accurate product details (vibrant red color), and the Style Trajectory captures more granular linguistic characteristics such as \textit{emphatic}, which results in the final better personalized text generation. 

\vspace{-0.5ex}
\section{Related Work} \label{sec:related-work}
\paragraph{LLM Personalization.} LLM personalization methodologies generally diverge into two directions. First, personalized preference optimization~\cite{li2024personalizedlanguagemodelingpersonalized, jang2023personalizedsoupspersonalizedlarge, wu2023finegrainedhumanfeedbackgives} focuses on learning specific user embeddings or fine-tuning parameters to align the model with user behavior. To inject personalized preference, instead of %finetuning the LLM to align 
aligning 
with general user preferences, researchers design multiple preference dimensions and personalize the reward weights~\cite{wu2023finegrainedhumanfeedbackgives} or preference assignment~\cite{jang2023personalizedsoupspersonalizedlarge}. More recent work focuses on extracting user embeddings directly from the data~\cite{li2024personalizedlanguagemodelingpersonalized} and learning customization reward functions~\cite{poddar2024personalizingreinforcementlearninghuman}.

Second, prompt-based personalization focuses on steering text generation by explicitly providing personal context within the input prompt~\cite{salemi2024lamplargelanguagemodels, kumar2024longlampbenchmarkpersonalizedlongform, au2025personalizedgraphbasedretrievallarge, ni2026grasper}. Early works focuses on retrieval personal context~\cite{salemi2024lamplargelanguagemodels, kumar2024longlampbenchmarkpersonalizedlongform}, while recent work explores the cold-start settings where users only entertain minimal histories~\cite{ni2026grasper, au2025personalizedgraphbasedretrievallarge}.

\vspace{-1ex}
\paragraph{Differential Rewards.} 

Traditional Reinforcement Learning from Human Feedback (RLHF)  optimizes a single scalar reward representing a ``universal'' preference, often suppressing minority viewpoints or conflicting objectives. Differential reward methods decompose this signal to capture fine-grained or group-specific preferences. Multi-Objective Reward Modeling uses separate reward heads for distinct attributes, enabling trade-offs or user-steerable inference~\cite{wu2023finegrainedhumanfeedbackgives, dong2023steerlmattributeconditionedsft}. Group-Conditional Rewards have been introduced to mitigate majority bias and address demographic or ideological diversity~\cite{bakker2022finetuninglanguagemodelsagreement}. 
Furthermore, Contrastive Reward frameworks isolate personal preferences by modeling the differential between a user-specific reward function and a general population baseline, enabling efficient adaptation without extensive retraining~\cite{poddar2024personalizingreinforcementlearninghuman, sorensen2024roadmappluralisticalignment, sorensen-etal-2025-value}. In a different domain, RAG-DDR~\cite{li2025ragddroptimizingretrievalaugmentedgeneration} proposes to optimize the model with signals from downstream generation, providing an alternative to typical reward-based learning paradigms.

\vspace{-0.5ex}
\section{Conclusion} \label{sec:conc}
In this paper, we presented PAT, a framework designed to tackle cold-start LLM personalization by decomposing user context into complementary writing-style and topic-knowledge trajectories. By leveraging an iterative dual-reasoning mechanism optimized via differential rewards, PAT effectively synthesizes noisy neighbor signals to align generation with user preferences, even when personal history is sparse. Our extensive experiments demonstrate that PAT significantly outperforms state-of-the-art baselines in sparse-data scenarios, achieving an average improvement of over 15\% for users with zero history. Our findings highlight the critical role of structured reasoning with differential data reward optimization in overcoming data sparsity.

\section*{Impact Statement}
This paper introduces PAT, a framework designed to significantly enhance the quality and alignment of personalized text generation, particularly in sparse-data scenarios where traditional methods fail. While our primary objective is to improve user experience in applications like recommendation systems and conversational agents, we acknowledge the potential for dual-use. The same multi-trajectory reasoning mechanisms that allow \method to effectively mimic specific writing styles and integrate topic-specific preferences could be exploited by malicious actors. Specifically, these capabilities could be utilized to generate highly convincing, personalized content for phishing attacks or social engineering campaigns, making them significantly more difficult for victims to detect than generic attempts. Consequently, as personalized generation technologies advance, it is imperative for the research community to develop robust countermeasures, such as detection frameworks or content watermarking, to mitigate these risks.

\bibliography{main}
\bibliographystyle{icml2025}

\newpage
\appendix
\onecolumn
\section{Additional Experimental Setup}
\label{app:exp-set-up}
In this section, we will expand the experiment section with additional experimental setups and details. First, we will introduce the datasets and their corresponding statistics. We will then have an extended discussion on the metrics.

\subsection{Datasets}
\label{appendices:dataset}
\begin{table}[h!]
\centering
\scriptsize
\resizebox{0.6\columnwidth}{!}{
\begin{tabular}{lccc}
\toprule
\textbf{Dataset} & \textbf{Train Size} & \textbf{Validation Size} & \textbf{Test Size} \\
\midrule
User-Product Review & 20,000 & 2,500 & 2,500 \\
Stylized Feedback & 20,000 & 2,500 & 2,500 \\
Hotel Experiences & 9,000 & 2,500 & 2,500 \\
\bottomrule
\end{tabular}}
\caption{Dataset split sizes across training, validation, and test sets for the four domains.}
\label{tab:split-stats}
\end{table}

We evaluate \method on the three benchmark datasets introduced in PGraphRAG benchmark~\cite{au2025personalizedgraphbasedretrievallarge}. These datasets cover diverse domains and graph structures, enabling us to assess the effectiveness of our method. 

\vspace{-1.75ex}
\paragraph{Amazon Review.}
The Amazon Review dataset is constructed from the Amazon Review 2023 corpus~\citep{hou2024bridging}. We build a user-item interaction graph where nodes represent users and products, and edges indicate review interactions between them. 

\vspace{-1.75ex}
\paragraph{Hotel Experience.}
The Hotel Experience dataset is collected from the Datafiniti Hotel Reviews dataset~\citep{au2025personalizedgraphbasedretrievallarge}. It contains user-hotel interaction data, where edges denote users' stays at hotels and are annotated with textual reviews. 

\vspace{-1.75ex}
\paragraph{Stylized Feedback Review.}
The Stylized Feedback Review dataset is derived from the Datafiniti Grammar and Online Product dataset~\citep{au2025personalizedgraphbasedretrievallarge}. It focuses on generating stylistic and domain-specific feedback from user-product interactions. This dataset emphasizes linguistic diversity and style adaptation.

Below we further provide the dataset statistics. In \cref{tab:split-stats}, we give the train/validation/test split statistics for the datasets. It is worth noting that the Hotel Experience dataset is a smaller dataset with a smaller training set, leading to the more inconsistent performance that we presented in the Experiment section. In \cref{tab:task-stats}, we introduce the task statistics for Long Text Generation and Short Text Generation. The datasets are constructed to reflect the real-world distribution~\citep{au2025personalizedgraphbasedretrievallarge}, which results in the sparse profiles as shown in the Average Profile Size. \method achieves more consistent and significant performance gain in scenarios where the output length is shorter, such as Short Text Generation and the User-Product Review (Amazon dataset), as longer text implicitly gives more context for text generation.
\begin{table*}[h!]
    \centering
    \scriptsize
    \begin{adjustbox}{width=1.0\textwidth}
        \begin{tabular}{lccccc} 
        \toprule
            \textbf{Task} & \textbf{Type} & \textbf{Avg. Input Length} & \textbf{Avg. Output Length} & \textbf{Avg. Profile Size} & \textbf{\# Classes} \\
        \midrule
            User-Product Review Generation & Long Text Generation & $3.754\pm2.71$ & $47.90\pm19.28$ & $1.05\pm0.31$ & - \\
            Hotel Experiences Generation & Long Text Generation & $4.29\pm2.57$ & $76.26\pm22.39$ & $1.14\pm0.61$ & - \\
            Stylized Feedback Generation & Long Text Generation & $3.35\pm2.02$ & $51.80\pm20.07$ & $1.09\pm0.47$ & - \\
        \midrule
            User-Product Review Title Generation & Short Text Generation & $30.34\pm37.95$ & $7.02\pm1.14$ & $1.05\pm0.31$ & - \\
            Hotel Experiences Summary Generation & Short Text Generation & $90.40\pm99.17$ & $7.64\pm0.92$ & $1.14\pm0.61$ & - \\
            Stylized Feedback Title Generation & Short Text Generation & $37.42\pm38.17$ & $7.16\pm1.11$ & $1.09\pm0.47$ & - \\
        \bottomrule
        \end{tabular}
    \end{adjustbox}
    \caption{
    Data statistics for the PGraphRAG Benchmark across the four datasets. For each task, we report the average input and output lengths (in words), measured on the test set using BM25-based retrieval with GPT. The average profile size indicates the number of reviews per user used for personalization.
    }
    % \caption{ \textcolor{blue}{Data statistics for PGraphRAG Benchmark across the four datasets. The table reports the average input length and average output length in words (done for the test set on \gpt on BM25 back on all methods). The average profile size for each task is the number of reviews a user has.}}
    \label{tab:task-stats}
\end{table*}

\vspace{-2ex}

\subsection{Tasks}
Here we present an extended discussion on the tasks that we used to evaluate \method: Long Text Generation and Short Text Generation. 

\vspace{-1.75ex}
\paragraph{Long Text Generation.}
The long text generation task focuses on producing detailed user reviews given a review title and the user's profile. The objective is to generate coherent and contextually relevant review text that aligns with the user’s preferences. This task evaluates the model’s capability for generating high-quality, personalized text.

\vspace{-1.75ex}
\paragraph{Short Text Generation.}
The short text generation task involves generating concise product titles or summaries given a user review. The challenge lies in distilling a longer text into a shorter title. This task assesses the model’s ability to distill information from highly personalized user context.

\subsection{Metrics}
For both long and short text generation tasks, we adopt widely used lexical overlap metrics, including ROUGE-1 and ROUGE-L, following prior work~\cite{au2025personalizedgraphbasedretrievallarge}. These metrics capture n-gram and subsequence overlaps between the generated output and ground-truth references. To complement these surface-level measures, we further incorporate LLM-as-a-Judge evaluation, where a strong language model provides comparative assessments of personalization and accuracy. We design the prompt based on prior studies which has been validated with human evaluators on the task of personalization~\citep{salemi2025reasoningenhancedselftraininglongformpersonalized}. The prompt for LLM-as-a-Judge evaluation is provided in Appendix \ref{app:lj}.

\subsection{Experimental Configuration}
All experiments were conducted on a single NVIDIA GeForce RTX 4090 GPU. Regarding the specific hyperparameters for our method, we set the maximum number of iterations $T = 10$. The retrieval and memory parameters were configured as $k_1 = k_2 = 5$ and $M_1 = M_2 = 3$. 

For our evaluation protocol, we first performed a hyperparameter search on the validation set. The configuration that yielded the best performance was then selected to conduct the final evaluation on the held-out test set.

\section{Aggregated Results}
In Table~\ref{tab:overall-results-avg-over-benchmarks}, we provide the aggregated results over all benchmarks. The table demonstrates that \method consistently outperforms the baseline methods (GraSPeR, PGraph, and LaMP) across the majority of metrics and tasks. Specifically, in the Long Text Generation task, our approach achieves the highest scores in Rouge-1, Rougle-L, and METEOR for both Qwen3 and LlaMA3 backbones, indicating superior content preservation and generation quality.

Similarly, in Short Text Generation, \method maintains a distinct advantage, particularly in LLM-based evaluation scores, where it surpasses the closest competitor by a notable margin (e.g., 3.733 vs. 3.653 with Qwen3). These results suggest that \method is robust and effective regardless of the output length or the underlying large language model used.

\section{\method Prompts}
\label{app:prompts}
In this section, we supply the prompts we used in \method. $\mathcal{P}_{style}$ is used in \cref{eq:style} to elicit style reasoning traces. $\mathcal{P}_{topic}$ is used in \cref{eq:knowledge} to elicit the topic reasoning traces. Lastly, $\mathcal{P}_{gen}$ is used in \cref{eq:gen} to generalize the final input for personalized text generation.

\begin{tcolorbox}[title=Style Reasoning Trace Prompt $\mathcal{P}_{style}$, colframe=gray]
\footnotesize   
\begin{verbatim}
System Prompt: You are a professional writing assistant whose task is to summarize 
the writing style of a user from the profile, which is past documents written by 
that user. The extracted writing style summary should contain the unique features 
of users writing style and preferences from the proile that are similar to the 
expected output.

Your task is to summarize the user writing style from the profile in a short bullet 
list (100 words or less). From the profile, you may infer the user's interests, 
preference, familiarity on various topics, etc. While inferring the user's interests, 
you can make reasonable guesses, e.g. people who are interested in topic A are also 
likely to be interested in topic B or if they write a sentence in a specific writing 
style on topic A it is likely they write it with the same style on topic B. As a 
concrete example, if a user writes "I am interested in action movies" in its past 
document, this is relevant to "I like to go to cinema". Another example would be if 
a person prefers specific words or phrases in their writing or using a specific 
grammar. You can also mention such words that they often use in your summary.

Your input:
- profile: the past documents written by the same person (might be empty). 

- similar profiles: past documents written by users that have similar writing style. 

Your output:
a list of bullet points and explanations describing writing style of the user. 

profile:
<Profile>

similar profiles:
<Similar Profiles>

Your output should be in the following format: 
Writing Style: <summarized writing style>. 

DO NOT include any other text in your output.
\end{verbatim}
   
\end{tcolorbox}

\begin{tcolorbox}[title=Topic Reasoning Trace Prompt $\mathcal{P}_{topic}$, colframe=gray]
\footnotesize
\begin{verbatim}
System Prompt: Your task is to summarize the product information from the product 
reviews in a short bullet list (100 words or less). From the review, you may summarize 
the key property of the product such as brands, makes, quality etc. While inferring 
the product summary, you can make reasonable guesses, e.g. a lot of people complains 
certain aspects of the products means there's certain key deficiency on the product. 
You also need to summarize the ratings given to the product. You are also welcome to 
make reasonable guesses, such as certain property would lead to low rating.

Your input:
- Product Reviews: the past reviews written on the product. 

Your output:
a list of bullet points and explanations describing the product and the rating it 
received.

profile:
<Product Text>

Your output should be in the following format: 
Product Summary: <summarized product summary>. 

DO NOT include any other text in your output.
\end{verbatim}
\end{tcolorbox}

\begin{tcolorbox}[title= Mutli-trajectory Generation Prompt $\mathcal{P}_{gen}$, colframe=gray]
\footnotesize
\begin{verbatim}
System Prompt: You are a personalized review generation assistant that generates 
high-quality reviews based on user history and context.

Instructions: Generate a review based on the specific writing style and product 
details provided below.

### Writing Style Context:
Original Example (Style Neighbor): {style_neighbor}
Style Summary: {style_summary}
### Product Context
Original Reviews (Product Neighbor): {product_neighbor}
Product Summary: {product_summary}
### Task
Review Title: {review_title}
Output Format:\nReview Text: <>
\end{verbatim}
\end{tcolorbox}

\section{LLM-as-a-Judge in Personalization}
\label{app:lj}
Traditionally, in the prior personalization benchmarks~\citep{au2025personalizedgraphbasedretrievallarge, kumar2024longlampbenchmarkpersonalizedlongform, salemi2024lamplargelanguagemodels}, personalized text generation has been evaluated with lexical overlap metrics such as ROUGE~\citep{lin-2004-rouge}. However, it has been shown that such metrics may fail to capture the semantic nuances and stylistic alignment in personalization. Thus, we adopt the LLM-as-a-Judge prompt from prior works on personalized text generation~\cite{salemi2025reasoningenhancedselftraininglongformpersonalized}, which is designed based on the evaluation paradigm introduced in \cite{liu-etal-2023-g}. Our prompt is introduced as follows.
\newpage 

\begin{tcolorbox}[title= LLM-as-a-Judge, colframe=gray]
\footnotesize
\begin{verbatim}
Please compare the generated text to the reference text based on how well they match 
and/or are similar.

Scoring Scale:
 1. Strongly disagree 
 2. Disagree
 3. Somewhat disagree
 4. Neither agree nor disagree
 5. Somewhat agree
 6. Agree 
 7. Strongly agree
 
Content to Evaluate: 
Reference Text (Ground Truth): {target_text}
Generated Text: {generated_text}

Provide only the numeric score (1–7).
\end{verbatim}
\end{tcolorbox}

We use Qwen2.5-7B as the judge LLM, and report the normalized score (0.1-0.7) in our main experiment table. ~\cite{salemi_reasoning-enhanced_2025} designed additional experiments to validate the effectiveness of the LLM-as-a-Judge evaluation. First, they conduct a human evaluation comparing 100 model outputs and find that the LLM-as-a-Judge scores agree with human preference in 73\% of cases, with a Pearson correlation of 0.46. Second, they design a controlled perturbation study by randomly replacing a portion of the personalized contexts with unrelated ones. The LLM-as-a-Judge scores decrease linearly as the perturbation rate increases, showing that the evaluator is sensitive to mismatched personalization.

While no automatic metric can fully replicate human evaluation for personalization—since the “true” judge of style and preference is the original user—LLM-as-a-Judge provides a scalable and semantically meaningful proxy. In our setting, it enables consistent evaluation across sparse and noisy contexts, capturing personalization quality beyond what lexical metrics can measure.

\begin{table}[t]
\renewcommand{\arraystretch}{0.5}
\setlength\tabcolsep{1.25pt}
\centering
\caption{Results averaged over all benchmarks.}
\label{tab:overall-results-avg-over-benchmarks}
\vspace{2mm}
\begin{tabular}{ccccccc}
\toprule
\textbf{Task} & \textbf{Metric} & \textbf{LLM} & 
\textbf{\method} (Ours) & 
\textbf{GraSPeR} & 
\textbf{PGraph} & \textbf{LaMP} \\
\midrule
\multirow{8}{*}{ \rotatebox{90}{\textcolor{googlegreen}{\textbf{\sc \bfseries Long Text Gen.}}} }
& \multirow{2}{*}{R-1 $\uparrow$}  & \textit{Qwen3}  & \textbf{0.231} & 0.199 & 0.217 & 0.175 \\[0.6ex]
&                           & \textit{LlaMA3}      & \textbf{0.235} & 0.190 & 0.208 & 0.189 \\
\cmidrule(l){2-7}
& \multirow{2}{*}{R-L $\uparrow$}  & \textit{Qwen3}             & \textbf{0.160} & 0.145 & 0.144 & 0.121 \\[0.6ex]
&              & \textit{LlaMA3}               & \textbf{0.164} & 0.148 & 0.147 & 0.126 \\
\cmidrule(l){2-7}
& \multirow{2}{*}{MET $\uparrow$}  & \textit{Qwen3}             & \textbf{0.193} & 0.158 & 0.187 & 0.144 \\[0.6ex]
&                           & \textit{LlaMA3}               & \textbf{0.180} & 0.137 & 0.155 & 0.154 \\
\cmidrule(l){2-7}
& \multirow{2}{*}{LLM $\uparrow$}  & \textit{Qwen3}  & 3.409 & 2.917 & \textbf{3.528} & 3.064 \\[0.6ex]
&              & 
\textit{LlaMA3} & \textbf{3.167} & 2.819 & 2.842 & 3.075 \\
\midrule
\multirow{8}{*}{\rotatebox{90}{\textcolor{googleblue}{\textbf{\sc \bfseries Short Text Gen.}}}}
& \multirow{2}{*}{R-1 $\uparrow$}  & \textit{Qwen3}             & \textbf{0.167} & 0.159 & 0.124 & 0.111 \\[0.6ex]
&                           & \textit{LlaMA3}               & \textbf{0.147} & 0.136 & 0.128 & 0.113 \\
\cmidrule(l){2-7}
& \multirow{2}{*}{R-L $\uparrow$} & \textit{Qwen3}             & \textbf{0.161} & 0.156 & 0.119 & 0.105 \\[0.6ex]
&                           & \textit{LlaMA3}               & \textbf{0.144} & 0.134 & 0.122 & 0.107 \\
\cmidrule(l){2-7}
& \multirow{2}{*}{MET $\uparrow$} & \textit{Qwen3}             & \textbf{0.149} & 0.126 & 0.124 & 0.113 \\[0.6ex]
&                           & \textit{LlaMA3}               & 0.106 & 0.121 & \textbf{0.124} & 0.120 \\
\cmidrule(l){2-7}
& \multirow{2}{*}{LLM $\uparrow$}   & \textit{Qwen3}  & \textbf{3.733} & 3.321 & 3.653 & 3.443 \\[0.6ex]
&              
& \textit{LlaMA3}               & \textbf{3.349} & 3.158 & 3.138 & 3.007 \\
\bottomrule
\end{tabular}
\vspace{-0.5em}
\end{table}

\section{Notations}
\label{sec:notation}
\begin{table*}[t]
\centering
\scriptsize
\setlength{\tabcolsep}{6pt}
\renewcommand{\arraystretch}{1.15}
\caption{Summary of key notation.}
\label{tab:notation}
\vspace{2mm}
\begin{tabular}{l p{0.78\linewidth}}
\toprule
% \textbf{Symbol} & \textbf{Definition} \\
% \midrule
$u$ & User. \\
$t$ & Topics. \\
$t_{\text{target}}$ & Target topic for personalized generation. \\
$x_{\text{target}}$ & Input corresponding to the target generation request. \\
$y$ & Ground-truth target text. \\
$\hat{y}$ & Generated personalized output text. \\
$\mathcal{H}_u$ & Historical profile of user $u$, consisting of past entries $(x_{u,n}, y_{u,n})$. \\
$(x_{u,n}, y_{u,n})$ & The $n$-th historical input-text pair for user $u$. \\
$\mathcal{A}_u$ & Auxiliary information for user $u$, defined as $(\mathcal{C}^{style}_u, \mathcal{C}^{topic}_u)$. \\
$\mathcal{C}^{style}_u$ & Style context capturing stylistic cues relevant to user $u$. \\
$\mathcal{C}^{topic}_u$ & Topic knowledge context capturing topic-specific information. \\
\midrule
$\mathcal{G}=(\mathcal{V},\mathcal{E})$ & User-topic bipartite graph. \\
$\mathcal{V}$ & Vertex set of the graph, $\mathcal{U}\cup\mathcal{T}$. \\
$\mathcal{E}$ & Edge set of the graph. \\
$\mathcal{U}$ & Set of user nodes. \\
$\mathcal{T}$ & Set of topic nodes. \\
$e_{u,t}$ &  Edge between user $u$ and topic $t$. \\
\midrule
$\phi(\cdot)$ & Style encoder mapping text to a continuous embedding. \\
$\mathbf{h}_u^{(0)}$ & Initial user embedding. \\
$\mathbf{h}_t^{(0)}$ & Initial topic embedding. \\
$\mathcal{R}_t$ & Set of all review texts associated with topic $t$. \\
$\mathbf{h}_v^{(k)}$ & Representation of node $v$ at GraphSAGE layer $k$. \\
$\mathbf{h}_u$ & Final user embedding after $K$ GraphSAGE layers. \\
$L$ & Number of GraphSAGE message-passing layers. \\
\midrule
$\mathcal{N}^{style}_u$ & Set of stylistically similar users retrieved for user $u$. \\
$\mathcal{N}^{topic}_u$ & Set of topic-aligned neighbors retrieved for user $u$. \\
$k_1$ & Number of stylistic neighbors retained. \\
$k_2$ & Number of topic-knowledge neighbors retained. \\
\midrule
$\pi_{\theta_s}$ & Style trajectory reasoning agent parameterized by $\theta_s$. \\
$\pi_{\theta_t}$ & Topic knowledge reasoning agent parameterized by $\theta_t$. \\
$\pi_{\theta_g}$ & Generation model parameterized by $\theta_g$. \\
$\mathcal{P}_{\text{style}}$ & Prompt construction function for the style reasoning agent. \\
$\mathcal{P}_{\text{topic}}$ & Prompt construction function for the topic reasoning agent. \\
$\mathcal{P}_{\text{gen}}$ & Prompt construction function for the final personalized text generation. \\
$s_u$ & Style reasoning trace produced by the style reasoning agent. \\
$p_{u,t_{\text{target}}}$ & Topic knowledge reasoning trace for user $u$ and target topic $t_{\text{target}}$. \\
\midrule
$M_1$ & Number of sampled style summaries. \\
$M_2$ & Number of sampled topic summaries. \\
$s_u^{(m)}$ & The $m$-th sampled style reasoning trace. \\
$p_u^{(m)}$ & The $m$-th sampled topic reasoning trace. \\
$\hat{y}^{(m)}$ & Generated output conditioned on a sampled trajectory. \\
$R(\cdot,\cdot)$ & Downstream task reward function. \\
$R^{(m)}$ & Reward assigned to the $m$-th generated output. \\
\midrule
$\theta_s,\theta_t,\theta_g$ & Parameters of the style reasoning agent, topic reasoning agent, and generator, respectively. \\
\bottomrule
\end{tabular}
\end{table*}

\end{document}